%% file: main.tex
\documentclass[sigconf]{acmart}
\usepackage{multirow}
\usepackage{pifont}
\usepackage{graphicx}
\usepackage{subcaption}
\usepackage{textcomp}
\usepackage{enumitem}
\usepackage{booktabs}
\usepackage{hyperref}
\AtBeginDocument{%
  }

\copyrightyear{2025}
\acmYear{2025}
\setcopyright{acmlicensed}
\acmConference[MM '25] {Proceedings of the 33rd ACM International Conference on Multimedia}{October 27--31, 2025}{Dublin, Ireland.}
\acmBooktitle{Proceedings of the 33rd ACM International Conference on Multimedia (MM '25), October 27--31, 2025, Dublin, Ireland}
\acmISBN{979-8-4007-2035-2/2025/10}
\acmDOI{10.1145/3746027.3755463}
\begin{document}
\title{Robust Understanding of Human-Robot Social Interactions through Multimodal Distillation}

\author{Tongfei Bian}
\orcid{0000-0001-5944-4157}
\email{t.bian.1@research.gla.ac.uk}
\affiliation{%
  \institution{University of Glasgow}
  \department{School of Computer Science}
  \streetaddress{18 Lilybank Gardens}
  \city{Glasgow}
  \country{United Kingdom}
  \postcode{G12 8RZ}
}

\author{Mathieu Chollet}
\orcid{0000-0001-9858-6844}
\email{mathieu.chollet@glasgow.ac.uk}
\affiliation{%
  \institution{University of Glasgow}
  \department{School of Computer Science}
  \streetaddress{18 Lilybank Gardens}
  \city{Glasgow}
  \country{United Kingdom}
  \postcode{G12 8RZ}
}

\author{Tanaya Guha}
\orcid{0000-0003-2167-4891}
\email{tanaya.guha@glasgow.ac.uk}
\affiliation{%
  \institution{University of Glasgow}
  \department{School of Computer Science}
  \streetaddress{18 Lilybank Gardens}
  \city{Glasgow}
  \country{United Kingdom}
  \postcode{G12 8RZ}
}

\begin{abstract}
There is a growing need for social robots and intelligent agents that can effectively interact with and support users. For the interactions to be seamless, the agents need to analyse social scenes and behavioural cues from their (robot's) perspective. Works that model human-agent interactions in social situations are few; and even those existing ones are computationally too intensive to be deployed in real time or perform  poorly in real-world scenarios when only limited information is available. We propose a knowledge distillation framework that models social interactions through various multimodal cues, and yet is robust against incomplete and noisy information during inference. We train a teacher model with multimodal input (body, face and hand gestures, gaze, raw images) that transfers knowledge to a student model which relies solely on body pose. Extensive experiments on two publicly available human-robot interaction datasets demonstrate that our student model achieves an average accuracy gain of 14.75\% over competitive baselines on multiple downstream social understanding tasks, even with up to 51\% of its input being corrupted. The student model is also highly efficient - less than $1$\% in size of the teacher model in terms of parameters and its latency is 11.9\% of the teacher model. Our code and related data are available at \href{github.com/biantongfei/SocialEgoMobile}{\textbf{github.com/biantongfei/SocialEgoMobile}}.
\end{abstract}



\begin{teaserfigure}
  \centering
  \includegraphics[width=0.9\textwidth]{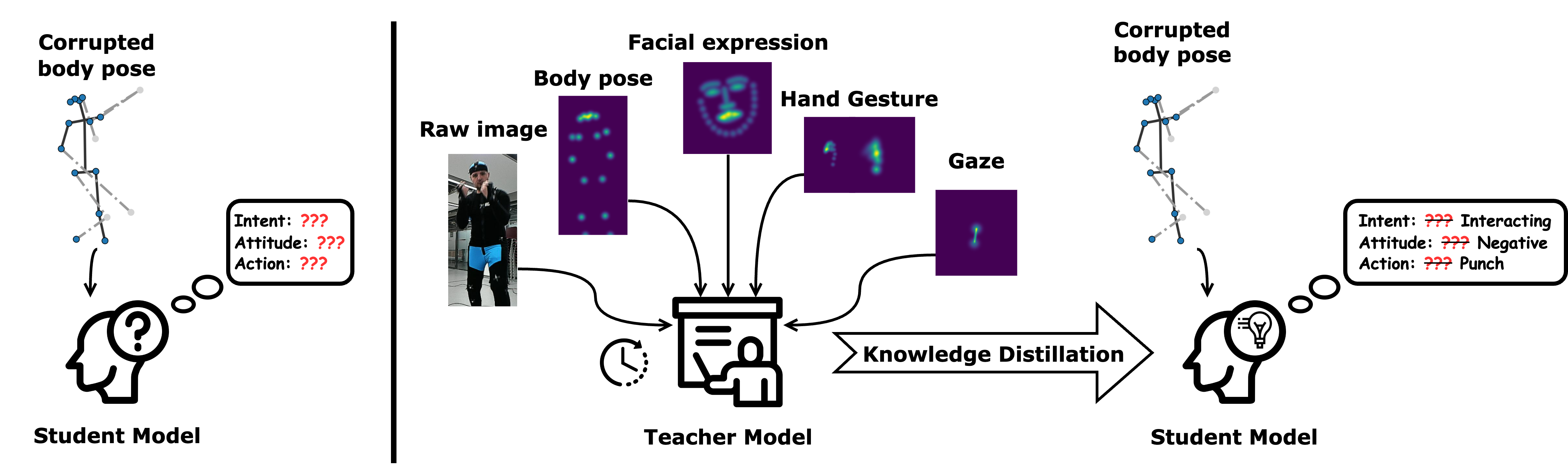}
  \vspace{-2mm}
  \caption{We aim to build a lightweight and robust model to understand human-robot social interactions. A novel knowledge distillation framework is built, where the teacher model uses multimodal cues (body pose, face and hand gestures, gaze and raw images) to learn a rich social representation. Through distillation, the lightweight student model that relies solely on corrupted body pose features learns multimodal social knowledge and is robust against incomplete and noisy inputs.}
  \label{fig:teaser}
\end{teaserfigure}


% \received[accepted]{5 July 2025}

\maketitle

\input{1_introduction/main}
\input{2_literature/main}
\input{3_method/main}
\input{4_experiments/main}
%
%\vspace{-2mm}
\section{Conclusion}
\label{sec:conclusion}
In this paper, we address the challenges of performance, robustness, and computational efficiency in forecasting users' social interaction intentions and behaviours from an egocentric perspective on mobile platforms or embedding systems. We propose SocialEgoMobile, a lightweight social understanding model based on knowledge distillation. In our knowledge distillation framework, the teacher model, SocialC3D uses multimodal inputs, including body poses, raw images, facial expressions, hand gestures, and gaze patterns, while the student model, SocialEgoMobile only relies on deliberately corrupted body pose features. By distilling knowledge from the teacher model’s integrated multimodal social representations and downstream task supervision, the student model learns to produce robust predictions under noisy and incomplete input conditions. This work offers an effective and lightweight solution for real-time social understanding in noisy environments, particularly suited for resource-constrained platforms such as social robots and intelligence social agents. Future work will focus on further enhancing the model’s robustness through real-world deployment and investigating its generalization to complex scenarios, including domain adaptation across viewpoints and environments.

% \begin{acks}
% This paper has been submitted to ACM Multimedia 2025.
% \end{acks}
\clearpage
\bibliographystyle{ACM-Reference-Format}
\bibliography{references}

\end{document}

%% file: 1_introduction/main.tex
\section{Introduction}
\label{sec:intro}
Recent application advances in social robotics and agents include personal care, navigation, and user interaction \cite{zhou2022human,zachiotis2018survey,blair2025blessing}. A central challenge in social AI and human-robot interaction (HRI) is enabling the systems to understand and forecast human social behaviours from an egocentric perspective, such as intent to interact, attitudes, and social actions \cite{bian2024interact}. This capability is particularly critical during the early stages of interaction, where understanding the user’s interaction intent allows the system to proactively prepare for natural and timely interactions. Early prediction enhances user experience and supports adaptive, context-aware responses \cite{avelino2021break}. However, deploying such predictive models in real-world settings remains challenging due to constraints in computational resources, input quality, and the need for real-time processing.
 
Understanding human social behaviours relies on multiple non-verbal signals \cite{urakami2023nonverbal}, each providing unique and complementary information into social interactions \cite{yumak2015multimodal}. Despite progress in modelling multimodal cues, several challenges remain in using them for social understanding in real-world deployment. \textbf{(1) Multimodal Integration:} Social understanding needs fusing information from diverse modalities that sometimes are inconsistent. Models need to integrate information and differences across modalities to extract coherent semantic representations. \textbf{(2) Real-World Challenges:} Egocentric perspectives introduce challenges such as limited field of view, camera motion and occlusions, which degrade the reliability of social understanding. \textbf{(3) Computational Efficiency:} Using multiple modalities requires computationally intensive models and more feature estimation modules, posing difficulties for real-time inference on resource-constrained systems, such as social robots.

To tackle these challenges, we propose a knowledge distillation framework that enables a lightweight student model to infer rich social cues from corrupted body pose input and jointly forecast three key aspects of social understanding: intent to interact, attitudes, and social actions, as shown in Fig.\ref{fig:teaser}. Our approach consists of two stages: first, we train a teacher model with multimodal inputs, including body, face and hand gestures, gaze and raw images, to learn high-quality social representations. Then, we distil this knowledge into a lightweight student model that relies solely on body pose while containing spatiotemporal noise to simulate real-world data degradation, ensuring robust performance under noisy conditions. Body pose features are crucial cues for social understanding \cite{bian2024interact} and rely only on lightweight feature extraction and analysis models.

In summary, our work enhances the robustness and efficiency of egocentric social understanding through the contributions:

\begin{itemize}[leftmargin=1em, topsep=0.3em]
    \item [1] \textbf{Knowledge distillation framework:} We propose a knowledge distillation framework where a teacher model SocialC3D using body, face and hand gestures, gaze and raw images, supervises a lightweight student model SocialEgoMobile that only uses body pose features. The average accuracy of the student model improved by 9.13\% over independent training on JPL-Social \cite{bian2024interact} and 11.32\% on the HARPER \cite{avogaro2024exploring}.
    \item [2] \textbf{Robustness to noises:} To enhance robustness against noisy and incomplete input, we introduce spatiotemporal corruption for the student model during distillation. When 51\% of spatiotemporal information is lost, the student maintains an average accuracy of 81.35\% on JPL-Social \cite{bian2024interact} (14.68\% improvement over independent training) and 59.05\% on HARPER \cite{avogaro2024exploring} (14.82\% improvement).
    \item [3] \textbf{Efficiency and deployment readiness:} The proposed student model SocialEgoMobile demonstrates high computational efficiency, requiring only 1\% of model parameters and 11.9\% latency compared to the teacher model SocialC3D.
\end{itemize}

%% file: 2_literature/main.tex
\section{Related Work}
\label{sec:liter}

\subsection{Social Intent Prediction}
Egocentric social intent prediction is crucial for social robots to establish natural, harmonious, and efficient interactions with users. A precise understanding of the user's social behaviour enables robots to timely adjust response strategies to meet user needs. Current research primarily models social intent as a binary classification task, focusing on predicting whether a user intends to interact with an intelligent agent. Approaches include analyzing motor cues such as body orientation, movement direction, and speed \cite{abbate2024self}, targeting users interested in interaction using visual and audible features \cite{pourmehr2017robust}, and using gaze information in virtual environments to infer interaction intent \cite{david2021towards}. However, these methods are limited to distinguishing between \textit{`interact'} and \textit{`not interact'}, failing to capture the richness and complexity of users' social intentions.

Human social intent estimation is inherently more nuanced than binary classifications and includes wider aspects. Recent research has explored various dimensions of users' social behaviours. Convolutional neural networks (CNNs) have been used to analyze facial expressions and body pose for attitude and emotion perception \cite{ilyas2021deep}, while multilayer perceptions (MLPs) have been utilized to process facial and thermal images to infer user emotions during HRI \cite{filippini2021facilitating}. Gated recurrent units (GRUs) capture the temporal dependencies of facial, acoustic, and textual features, enabling the detection of dynamic changes in users' attitudes and emotions during conversations \cite{li2020attention}. Similarly, interaction action prediction has gained attention. Using cascade histogram of time series gradients can capture past spatio-temporal features and predict users' future social actions \cite{ryoo2015robot}. Social action and wider social signal in multi-agent scenarios have also been explored, by integrating body, facial, and hand pose features to capture user's body movement \cite{bian2024interact}.

\subsection{Non-Verbal Cues in Multimodal HRI}
In Human-Robot Interaction (HRI), accurately interpreting the user’s identity, social intent and behaviours requires integrating multimodal cues \cite{yumak2015multimodal}. Meanwhile, non-verbal cues play a crucial role in social understanding \cite{urakami2023nonverbal}. For instance, while detecting sarcasm through text is challenging, recognizing subtle body movements such as shoulder shrugs allows humans to accurately identify sarcasm \cite{castro2019towards}. Consequently, capturing and interpreting mutlimodal non-verbal cues in HRI is essential for assessing the users. Recent research has explored various strategies for multi-modal fusion to enhance the accuracy of interaction models. For example, combining RGB images, facial expressions, and distance information can effectively assess user engagement \cite{ben2019fly} and interest \cite{foster2019mummer} when interacting with service robots in public spaces. Facial expression and gaze information can be used to infer whether a user requires assistance in collaborative tasks \cite{wilson2022help}. To further improve robustness in real-world scenarios, researchers tried to model temporal information of non-verbal signals. Recurrent models have been used to model gestures and RGB video streams for identifying user action commands \cite{liu2018towards}, and to detect user emotions during collaboration by analyzing body movements and facial expressions \cite{paez2022human}. Pose skeletons and RGB information have been fused to predict interaction actions in single-user \cite{islam2020hamlet} and multi-user scenarios \cite{yasar2024imprint}. Non-verbal modalities including body, face, hand gestures, gaze and raw images provide rich contextual information that enhances a robot’s ability to accurately understand to users' intentions, attitudes, and actions, improving the effectiveness of HRI systems.

\vspace{2mm}

\noindent \textbf{Research gap.} While deep learning models using multimodal cues have advanced social understanding, their high computational demands \cite{liang2024foundations} and sensitivity to noisy or incomplete inputs \cite{drenkow2021systematic} limit deployment on resource-constrained platforms. Existing knowledge distillation approaches primarily aim to retain accuracy under ideal conditions \cite{arani2021noise}, neglecting robustness and integrating multimodal information using a single modality. To overcome these issues, we propose a robust distillation framework where a lightweight, body pose-only student model learns from a multimodal teacher, achieving efficient, resilient, and accurate social understanding.

%% file: 3_method/main.tex
\section{Methodology}
\label{sec:method}
\input{3_method/fig_distillation}

\input{3_method/problem_definition}

\input{3_method/teacher}

\input{3_method/student}

\input{3_method/knowledge_distillation}

\input{3_method/input_corruption}

%% file: 3_method/fig_distillation.tex
\begin{figure*}[ht]
  \centering
  \includegraphics[width=0.9\linewidth]{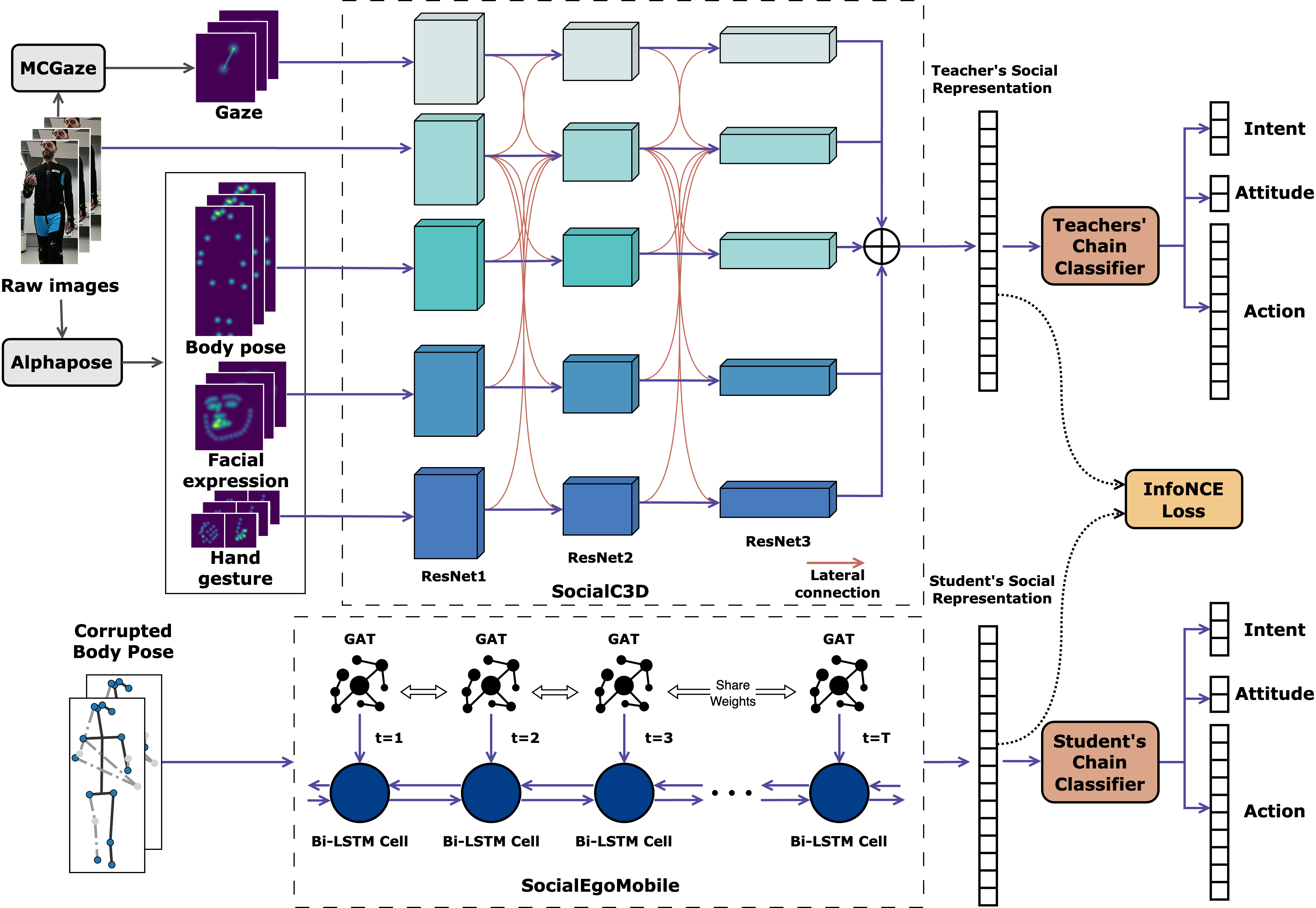}
  \vspace{-2mm}
  \caption{Our knowledge distillation framework uses SocialC3D as the teacher model, which fuses raw images, body, face, hand gestures, and gaze information, producing a multimodal social representation. The lightweight student model, SocialEgoMobile, uses only body pose features to output social representations. The framework maximises the mutual information of the social representations of the teacher and student model, to transfer social knowledge.}
  \label{fig:distillation}
  \vspace{-5mm}
\end{figure*}

%% file: 3_method/problem_definition.tex
\subsection{Problem Definition}
Given a sequence of egocentric videos captured by a social robot, we aim to forecast the user's social intent, attitude and action in real-time by observing the early stage (the first second). So we propose a knowledge distillation framework where a lightweight student model is trained to rely on body pose features, while a teacher model uses multimodal inputs (body, face, hand gesture, gaze and raw images), as shown in Fig.\ref{fig:distillation}. Body pose features, facial keypoints, and hand keypoints are extracted from the video using Alphapose \cite{fang2022alphapose} and gaze features are extracted using MCGaze \cite{guan2023end}. Formally, we define the input signals at step $t$ as follows:

\begin{itemize}[leftmargin=1em, topsep=0.3em]
    \item $\boldsymbol{x}_{t}^{\text{Pose}}$: The coordinates of 17 body pose keypoints
    \item $\boldsymbol{x}_{t}^{\text{Image}}$: The raw images
    \item $\boldsymbol{x}_{t}^{\text{Face}}$: The coordinates of 68 facial keypoints
    \item $\boldsymbol{x}_{t}^{\text{Hand}}$: The coordinates of 42 hands keypoints
    \item $\boldsymbol{x}_{t}^{\text{Gaze}}$: The coordinates of the beginning and end of gaze
\end{itemize}

The \textbf{teacher model} \( F_{teacher} \) is trained with the complete set of input modalities:
\begin{equation}
    R_{teacher} = F_{teacher}(\boldsymbol{x}_{t}^{\text{Pose}}, \boldsymbol{x}_{t}^{\text{Image}}, \boldsymbol{x}_{t}^{\text{Face}}, \boldsymbol{x}_{t}^{\text{Hand}}, \boldsymbol{x}_{t}^{\text{Gaze}}), t \in [0, T]
\end{equation}
where $R_{teacher}$ represents the high-level social representation outputed by the teacher model and $T$ represents the length of the input sequence. The lightweight student model $F_{student}$ relies solely on body pose features $\boldsymbol{x}_{t}^{\text{Pose}}$:
\begin{equation}
    R_{student} = F_{student}(\boldsymbol{x}_{t}^{\text{Pose}}), t \in [0, T]
\end{equation}
where $R_{student}$ represents the social representation outputed by the student model.

For the subtasks of forecasting user's intent to interact, attitude, and social actions, the social representations $R$ extracted by the teacher or student model are processed by a \textbf{Chain} hierarchical classifier \cite{bian2024interact}, which simulates how human infer others' social intents, predicting the results of three downstream tasks jointly:
\begin{equation}
    \mathcal{L}_{intent},\mathcal{L}_{attitude},\mathcal{L}_{action}=Classifier(R)
\end{equation}

%% file: 3_method/teacher.tex
\subsection{Teacher Model}
To integrate multiple modalities, we employ a teacher model, SocialC3D which is based on PoseC3D \cite{duan2022revisiting}, a spatio-temporal feature extraction framework using 3D convolutional neural networks (3D-CNN). PoseC3D converts sequences of 2D pose keypoints into 3D heatmap stacks, according to their coordinates and confidence scores. The heatmaps are converted as follows:
\begin{align}
    H_i(x, y) = \exp\left( -\frac{(x - x_i)^2 + (y - y_i)^2}{2\sigma^2} \right) * c_i
\end{align}
where $\sigma$ is the Gaussian standard deviation of the heatmap; ($x$, $y$) are the position in the heatmap; ($x_i$, $y_i$) and $c_i$ are the coordinates and confidence scores of the keypoint $i$. This approach exhibits greater robustness to pose estimation error than skeleton-based methods \cite{duan2022revisiting}. Also, the lateral connections between the RGB and Pose pathway fuse information on multiple levels.

Based on this, we introduce SocialC3D to incorporate additional modalities, including facial expressions, hand gestures, and gaze, into 3D heatmap stacks as input, which are processed using the same ResNet structure \cite{he2016deep} as the body pose pathway. For gaze vector, we use the centre of the head bounding box as the beginning and calculate the end based on the gaze direction and the diagonal length of head bounding box. SocialC3D incorporate lateral connections to propagate intermediate representations across modalities at multiple stages, enabling deeper feature integration. After the ResNets, the outputs of different modalities are concatenated and passed through a linear layer to produce a 16-dimensional multimodal social representation.

%% file: 3_method/student.tex
\subsection{Student Model}
The student model takes only body pose features as input. Compared to raw visual information, body language provides a more direct and interpretable cue for social understanding and body pose is one of the most effective modalities. Additionally, body posture inherently contains some facial and hand dynamics, allowing the model to integrate multimodal cues through knowledge distillation. Moreover, body pose, as a high-level abstraction, supports lighter analysis models and pose estimation methods, enhancing the student model’s suitability for resource-constrained settings.

To achieve a balance between performance, robustness to noise and computational efficiency, we developed \textbf{SocialEgoMobile} as our student model, consisting of a two-layer graphical attention network (GAT) \cite{veličković2018graph} and a single-layer bi-directional Long Short Term Memory (Bi-LSTM) \cite{graves2005framewise}. By dynamically adjusting the attentional weights between neighbouring joints, GAT can capture the spatial relationships of the body pose skeleton for each frame, focusing on useful information and limiting the noise. Meanwhile, Bi-LSTM captures the temporal dynamic of the input sequence forward and backward and filters noise using forgetting gates. The hidden dimension of the GATs is 16, and that of the Bi-LSTM is 128. The outputs are projected into a 16-dimensional social representation via a linear layer. This lightweight yet effective design maintains stable performance under noisy input conditions, thus enhancing robustness and making it suitable for deployment in resource-limited environments. 

%% file: 3_method/knowledge_distillation.tex
\subsection{Knowledge Distillation}
To transfer social knowledge from the teacher model to the student model, we adopt a feature-based distillation method, InfoNCE (Information Noise Contrastive Estimation)  \cite{oord2018representation}. The student learns from the ground truth labels of the downstream tasks and the difference between its social representation and that of the teacher. InfoNCE estimates and maximises its mutual information by enhancing the similarity between the social representations of teacher and student models of the same video clip while reducing the similarity between different ones. The distillation loss is defined as:
\begin{align}
    {{Loss}_{\text{infoNCE}}} = - \mathop{{}\mathbb{E}}\left[\log \frac{\exp(f(x, x^{+}))}{\sum_{x_j \in X} \exp(f(x, x_j))}\right] \label{loss}
\end{align}
where $x$ is the teacher’s social representation for a given video clip, $x^{+}$ is the representation from the studen for the same clip, and $x_j$ is the representation from the student for other clips. The similarity function $f(x, x')$ is defined using cosine similarity as follows:
\begin{align}
    \mathbf{}{f}(\mathbf{x}, \mathbf{x'}) = \frac{\mathbf{x} \cdot \mathbf{x'}}{\|\mathbf{x}\| \|\mathbf{x'}\|}
\end{align}

%% file: 3_method/input_corruption.tex
\subsection{Input Corruption}
\label{3_input_corruption}
In real-world applications, social understanding models must work under the computational constraints of mobile platforms and be robust to incomplete and noisy inputs. To simulate such conditions, we deliberately introduce corruption in spatial and temporal dimensions. In the spatial domain, a fixed number of body pose keypoints are randomly corrupted by setting their coordinate values and confidence scores to zero or replacing the coordinates with normalized random values. This mimics localized body occlusions and pose estimation errors. In the temporal domain, we randomly corrupt frames by setting all keypoint values to zero or applying normalised random noise, simulating transient full-body occlusions. These corruptions force the student model to learn robust representations and maintain performance under incomplete and noisy input.

%% file: 4_experiments/main.tex
\section{Experiments}
\label{sec:experiments}

\input{4_experiments/data}

\input{4_experiments/experiment_setting}

\input{4_experiments/performance}

\input{4_experiments/robustness_analysis}

\input{4_experiments/ablation_experiments}

\input{4_experiments/key_takeaways}

%% file: 4_experiments/data.tex
\subsection{Data}

\input{4_experiments/fig_data_example}

To evaluate the performance and robustness of our models in egocentric social understanding, we use two datasets of human-robot social interactions recorded from the robot's perspective. Examples of the datasets are shown in Fig. \ref{fig:data_example}.

\textbf{JPL-Social} \cite{bian2024interact} is an extension of the JPL-Interaction datasets \cite{ryoo2013first,ryoo2015robot}, focused on the user's social actions. The original dataset captures 8 social actions performed by the users in the centre of the screen: \textit{handshake, hug, pet, wave, punch, throw, point,} and \textit{leave}. JPL-Social extends this by annotating additional users in the background during multi-person scenarios, introducing two additional action label: \textit{gaze} and \textit{no response}. Furthermore, JPL-Social provides annotations for social intent (categorized as \textit{interacting, interested} and \textit{not interested}) and attitude (\textit{positive} or \textit{negative}) for each user. The dataset comprises 290 interaction sequences involving 16 participants in diverse environmental settings.  

\textbf{HARPER} \cite{avogaro2024exploring} is a dataset recorded from multiple perspectives using five cameras mounted on a quadrupedal robot. We use 6 social actions relevant to robot-oriented interactions: \textit{crash, walk-avoid, touch, walk-stop, punch} and \textit{kick}, so there are 260 sequences. In addition, HARPER includes variations in viewing angle and resolution, increasing the difficulty of understanding users' behaviours. Following the protocol of JPL-Social \cite{bian2024interact}, we manually annotate HARPER for the intent and attitude prediction tasks.

%% file: 4_experiments/fig_data_example.tex
\begin{figure}
    \centering
    \includegraphics[width=0.9\linewidth]{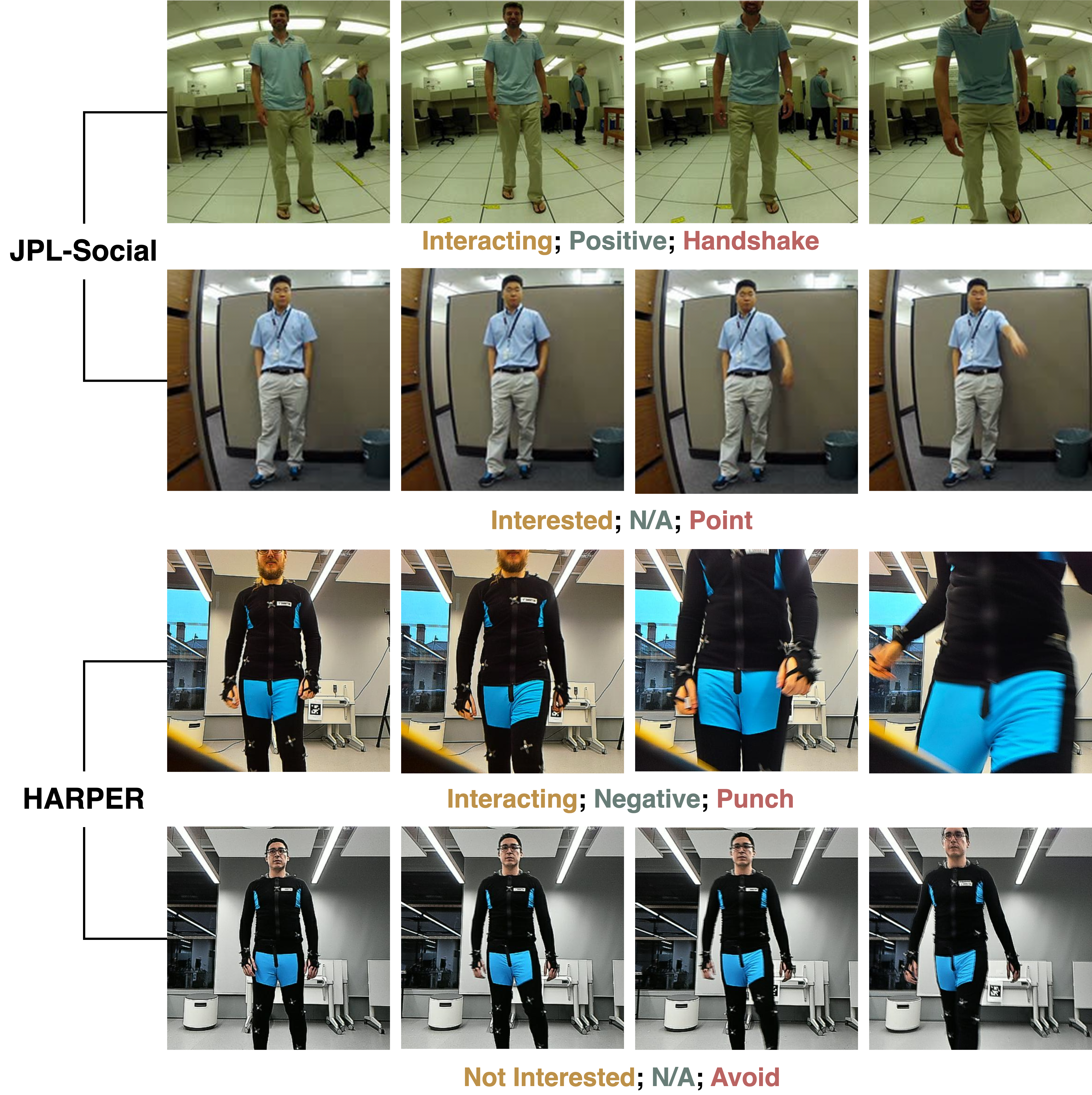}
    \vspace{-4mm}
    \caption{We set the FPS of two datasets to 10 and limited the observation window to the first second of social interactions.}
    \label{fig:data_example}
    \vspace{-6mm}
\end{figure}

%% file: 4_experiments/experiment_setting.tex
\subsection{Experiment Setting}
We set the Frames Per Second (FPS) of the videos to 10 and limited the observation window to the first second of social interaction, which starts when the user approaches the social agent, so the inputs have 10 frames. To expand the training set, we randomly cropped and horizontally flipped video frames,  augmenting the number of videos 20 times for richer training data. We train and test the models on the two datasets separately. Given our focus on model efficiency, we report the number of model parameters and the average latency for each input. All experiments were conducted on an NVIDIA RTX 4070 GPU.

All models are trained for 40 epochs with a batch size of 96. Training is early stopped if the average performance across the three subtasks does not improve for 5 epochs. Each modality channel in the SocialC3D is trained independently first and then jointly fine-tuning. To enhance performance, the body pose and raw image channels of all teacher models are initialized with pre-trained weights on Kinetics 400 \cite{kay2017kinetics}. SocialC3D is optimized using SGD \cite{sutskever2013importance}, while SocialEgoMobile uses the Adam optimizer \cite{kingma2014adam}. During knowledge distillation, the total loss is computed as a weighted sum of the classification loss and the distillation loss, with weights of 0.8 and 0.2. The temperature for the InfoNCE \cite{oord2018representation} is set to 0.1.

%% file: 4_experiments/performance.tex
\subsection{Performance of the Teacher Model}

\input{4_experiments/table_teacher}

As the teacher model, we compare SocialC3D with several state-of-the-art methods. Although originally designed for pose skeletons, these methods can also process facial and hand keypoints. To ensure a fair comparison, we incorporated the raw image and gaze information channels from SocialC3D into all teacher models, aligning the modalities across teacher models. In Table \ref{tab:teacher_models}, SocialC3D consistently achieves superior performance on three downstream social tasks on both datasets. Compared with the second-best model, MS-G3D \cite{liu2020disentangling}, SocialC3D achieves an average accuracy improvement of 5.85\% on JPL-Social and 5.77\% on HARPER, which highlights the effectiveness of SocialC3D in understanding social cues and forecasting users' social intentions from an egocentric perspective. We attribute this improvement to two key factors. First, instead of directly using pose features, converting them into heatmaps enhances robustness to pose estimation errors, which is particularly pronounced in egocentric settings due to narrow and jittery fields of view. Second, our experiments focus on the first 10 frames of social interactions. 3D CNNs are suitable for capturing short-term temporal dynamics, making them more appropriate for our experiment setting. In contrast, models like ST-GCN \cite{yan2018spatial} and MS-G3D \cite{liu2020disentangling} rely on complete and accurate pose features, making them more vulnerable to missing joints and estimation errors. ST-TR \cite{plizzari2021spatial}, while effective in modelling long-term dependencies due to using transformers \cite{vaswani2017attention}, struggles with short sequences. Although SocialEgoNet \cite{bian2024interact} is designed to capture users’ non-verbal social cues from an egocentric perspective, its lightweight architecture limits its capacity to extract fine-grained information.

In addition, we report the performance of SocialEgoMobile using only body pose to compare its accuracy and computational efficiency against the teacher models. Compared to SocialC3D, SocialEgoMobile's average accuracy of downstream tasks was 13.89\% lower on JPL-Social and 33.97\% lower on HARPER. The larger performance gap on HARPER is attributed to its recording setup, where the robot dog has a lower viewing angle and fewer frames with full-body visibility, resulting in noisier and more incomplete pose inputs, as shown in Fig.\ref{fig:data_example}. In such conditions, high-level features such as body pose become less reliable. These features rely on per-frame feature estimation methods, which lack temporal context and will produce unstable outputs, leading to fragmented feature sequences and degraded model understanding. In contrast, raw image sequences provide a temporally coherent context. Thus, teacher models with image inputs can maintain high performance on HARPER, while SocialEgoMobile, relying solely on body pose, suffers from significant degradation. In terms of computational efficiency, we report the number of parameters and the inference latency for all methods. Since the pose and gaze estimation algorithms can run at 10 FPS, the reported latency only includes the model inference time and the feature extraction time for the last frame. SocialEgoMobile only needs 0.43 M model parameters (0.98\% of SocialC3D) and 3.25 ms latency (11.9\% of SocialC3D). These results show that SocialEgoMobile is lightweight enough for real-time social understanding on mobile platforms with limited resources.

%% file: 4_experiments/table_teacher.tex
\begin{table*}[t]
\centering
\caption{Comparison of SocialC3D and SocialEgoMobile with state-of-the-art. '+' indicates that the model uses additional modalities same as SocialC3D. $\Delta_1$ denotes the time to extract whole-body pose and gaze features from a single frame, which is 4.96 ms under our experimental setup. $\Delta_2$ refers to the extraction time for body pose features, which is 3.06 ms.}
% \vspace{-2mm}
\renewcommand*{\arraystretch}{1.2}
\small
\resizebox{0.95\textwidth}{!}{
\begin{tabular}{l|cc|cccccc|cccccc}
\toprule
\multirow{3}{*}{\textbf{Model}} &
  \multirow{2}{*}{\textbf{Params}} &
  \multirow{2}{*}{\textbf{Latency}} &
  \multicolumn{6}{c}{\textbf{JPL-Social}} &
  \multicolumn{6}{c}{\textbf{HARPER}}\\
  \phantom{0} &
  \phantom{0} &
  \phantom{0} &
  \multicolumn{2}{c}{\textbf{Intent}} &
  \multicolumn{2}{c}{\textbf{Attitude}} &
  \multicolumn{2}{c}{\textbf{Action}} &
  \multicolumn{2}{c}{\textbf{Intent}} &
  \multicolumn{2}{c}{\textbf{Attitude}} &
  \multicolumn{2}{c}{\textbf{Action}} \\
  \phantom{0} &
  (M) $\downarrow$ &
  (ms) $\downarrow$ &
  \textbf{Acc.} &
  \textbf{F1} & 
  \textbf{Acc.} &
  \textbf{F1} & 
  \textbf{Acc.} &
  \textbf{F1} & 
  \textbf{Acc.} &
  \textbf{F1} & 
  \textbf{Acc.} &
  \textbf{F1} & 
  \textbf{Acc.} &
  \textbf{F1} \\
  \midrule
ST-GCN\textsuperscript{+} \cite{yan2018spatial}&
  43.86 &
  $\Delta_1$ + 10.31 &
  86.90 &
  85.93 &
  76.19 &
  74.56 &
  71.43 &
  65.23 &
  86.54 &
  84.02 &
  73.08 &
  74.68 &
  78.85 &
  77.18 \\
ST-TR\textsuperscript{+} \cite{plizzari2021spatial}&
  58.48 &
  $\Delta_1$ + 14.33 &
  79.76 &
  77.82 &
  59.52 &
  58.87 &
  48.81 &
  46.76 &
  75.00 &
  66.37 &
  65.38 &
  58.91 &
  59.61 &
  39.63 \\
MS-G3D\textsuperscript{+} \cite{liu2020disentangling}&
  48.82 &
  $\Delta_1$ + 13.74 &
  88.10 &
  88.11 &
  80.95 &
  78.12 &
  76.19 &
  74.59 &
  90.38 &
  88.11 &
  78.85 &
  76.52 &
  80.77 &
  74.00 \\
SocialEgoNet\textsuperscript{+} \cite{bian2024interact}&
  37.78 &
  $\Delta_1$ + 10.05 &
  86.90 &
  86.54 &
  77.38 &
  75.68 &
  71.43 &
  68.93 &
  86.54 &
  77.29 &
  75.00 &
  73.72 &
  78.85 &
  71.78 \\
\textbf{SocialC3D}&
  48.49 &
  $\Delta_1$ + 22.34 &
  \textbf{92.85} &
  \textbf{91.25} &
  \textbf{88.10} &
  \textbf{88.91} &
  \textbf{82.14} &
  \textbf{78.09} &
  \textbf{96.15} &
  \textbf{96.14} &
  \textbf{82.69} &
  \textbf{88.28} &
  \textbf{88.46} &
  \textbf{88.35}\\ 
  \midrule
SocialEgoMobile&
  \textbf{0.43} &
  \textbf{$\Delta_2$ + 0.19} &
  86.90 &
  85.26 &
  71.43 &
  68.61 &
  67.86 &
  62.63 &
  69.23 &
  66.92  &
  38.46 &
  42.07 &
  51.92 &
  46.56 \\ 
  \bottomrule
\end{tabular}
}
\label{tab:teacher_models}
\vspace{-2mm}
\end{table*}

%% file: 4_experiments/robustness_analysis.tex
\subsection{Robustness Analysis on the Student Model}

\input{4_experiments/fig_corrupted_information}

To address the practical challenges faced by mobile socially intelligent agents from an egocentric perspective, such as occlusion of user body parts and pose estimation error, we deliberately introduced random spatiotemporal corruption into the input of the student model SocialEgoMobile, as mentioned in Section \ref{3_input_corruption}. The goal is to improve robustness by using the multimodal and uncorrupted knowledge extracted from the teacher model, SocialC3D, to supervise the learning of the student model with corrupted inputs. During the robustness experiments, student models were trained for 70 epochs. The first 30 epochs are the warm-up phase, where the corruption rate is gradually increased to the target value to prevent unstable training due to excessive information loss.

\input{4_experiments/fig_modal}

Fig.\ref{fig:corruption_inputs} compares SocialEgoMobile’s performance under independent learning and knowledge distillation. Under independent learning, performance degrades as corruption increases. When 30\% of joints and frames are corrupted, $30\% \oplus 30\% = 51\%$ of the information in the input was corrupted. This ratio maintains a balance of noise and useful information. The average accuracy across three subtasks drops by 8.85\%, demonstrating the increased difficulty and robustness demands. In contrast, with knowledge distillation, SocialEgoMobile consistently outperforms the independently trained counterpart across all input settings. When 30\% of joints and 30\% of frames are corrupted, SocialEgoMobile trained via distillation surpasses the performance of the independently trained one without input corruption. This suggests that knowledge distillation effectively transfers social knowledge from the teacher model, which incorporates additional modalities and complete inputs to the student model. So it can more accurately capture valuable social cues even when facing corrupted inputs. The performance gains are particularly significant for the user attitude and social action prediction tasks, which rely more on detailed and accurate input.

Whether trained independently or via knowledge distillation, SocialEgoMobile is more sensitive to the corruption of spatial information than temporal information. This is because temporal patterns in human behaviour tend to be more regular and predictable, allowing models to more effectively compensate for missing frames and filter out noise. In contrast, spatial patterns of pose skeletons are more complex, such as body structure and human kinematics.

%% file: 4_experiments/fig_corrupted_information.tex
\begin{figure*}[t]
    \centering
    \begin{subfigure}[b]{0.9\textwidth}
        \centering
        \includegraphics[width=\textwidth]{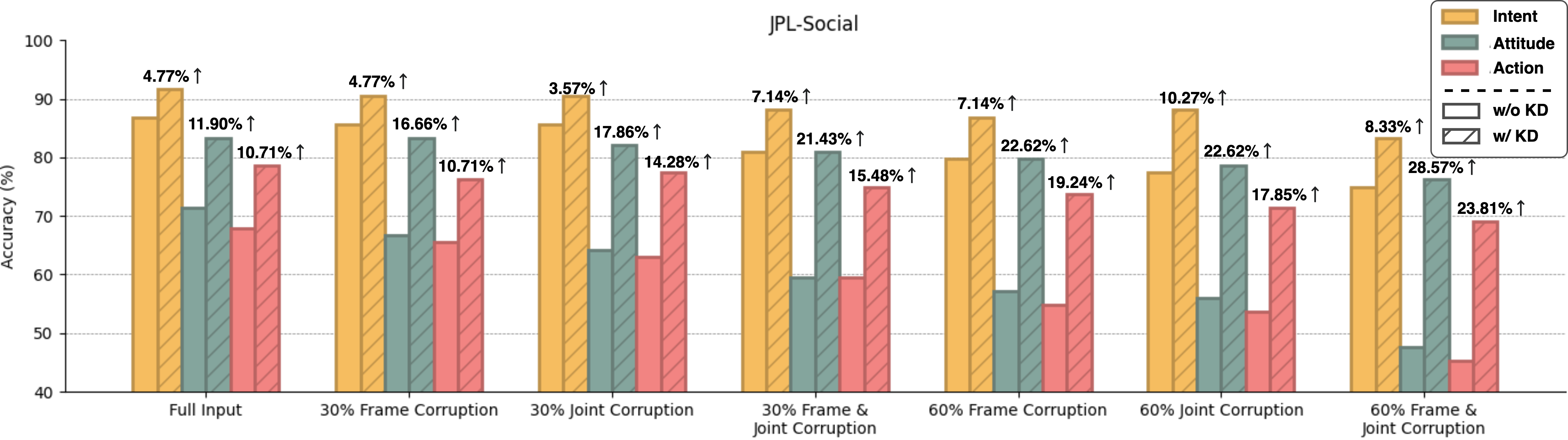}
        % \caption{JPL-Social}
    \end{subfigure}
    
    \vspace{2mm}
    
    \begin{subfigure}[b]{0.9\textwidth}
        \centering
        \includegraphics[width=\textwidth]{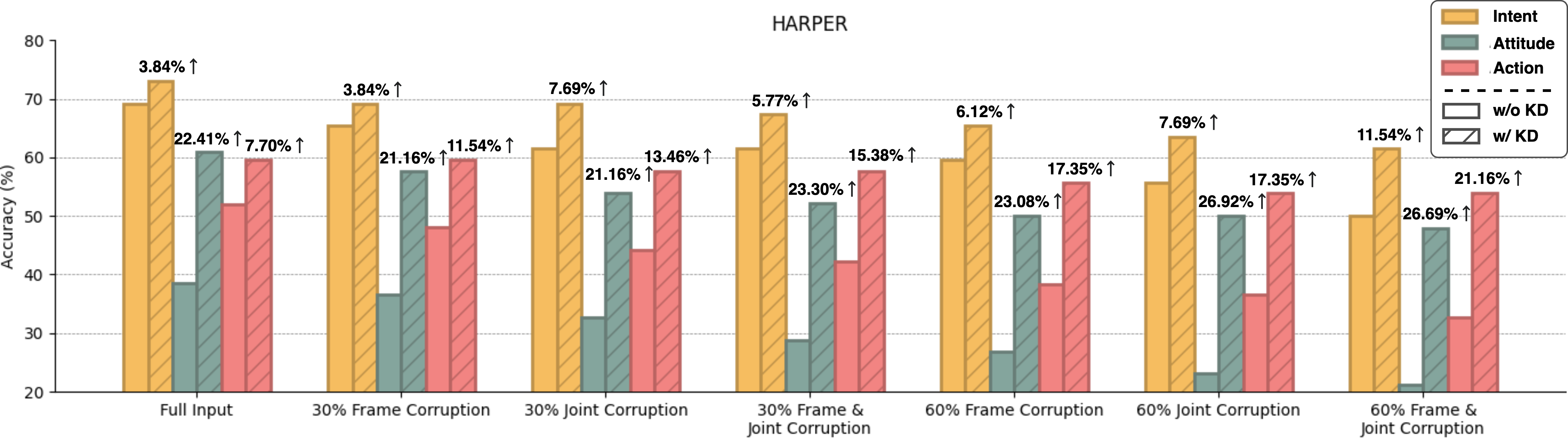}
        % \caption{HARPER}
    \end{subfigure}
    \vspace{-2mm}
    \caption{Knowledge distillation (KD) consistently improves the performance of the student model, SocialEgoMobile, under spatio-temporal corruption on all three downstream tasks, interaction intent, attitude, and social action forecast. Improvements on downstream task accuracy through distillation are labelled.}
    \label{fig:corruption_inputs}
    \vspace{-2mm}
\end{figure*}

%% file: 4_experiments/fig_modal.tex
\begin{figure*}[t]
  \centering
  \begin{minipage}{0.9\textwidth}
  \begin{subfigure}{0.5\textwidth}
  \centering
    \includegraphics[width=\linewidth]{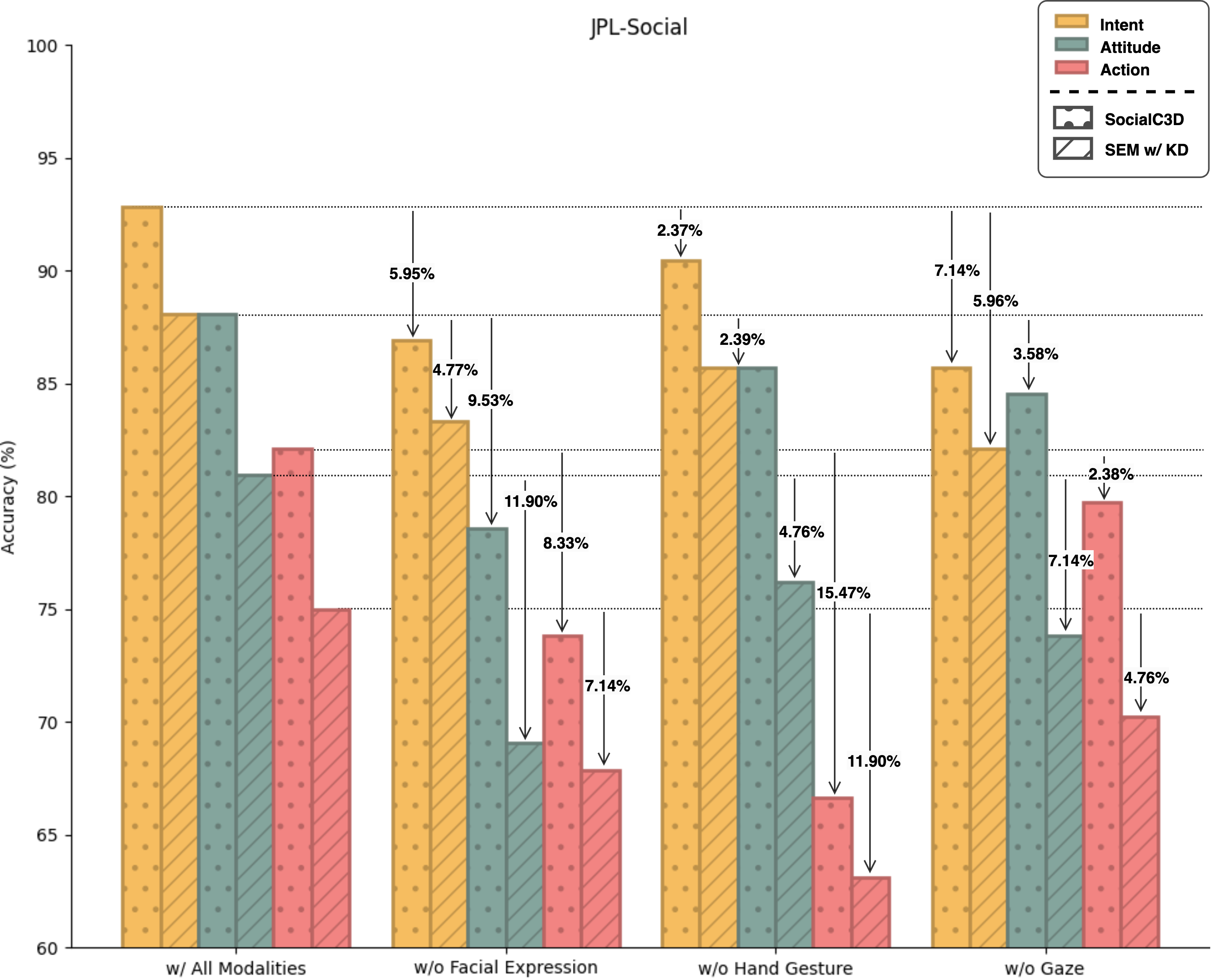}
    % \caption{JPL-Social}
  \end{subfigure}
  \hfill
  \begin{subfigure}{0.5\textwidth}
  \centering
    \includegraphics[width=\linewidth]{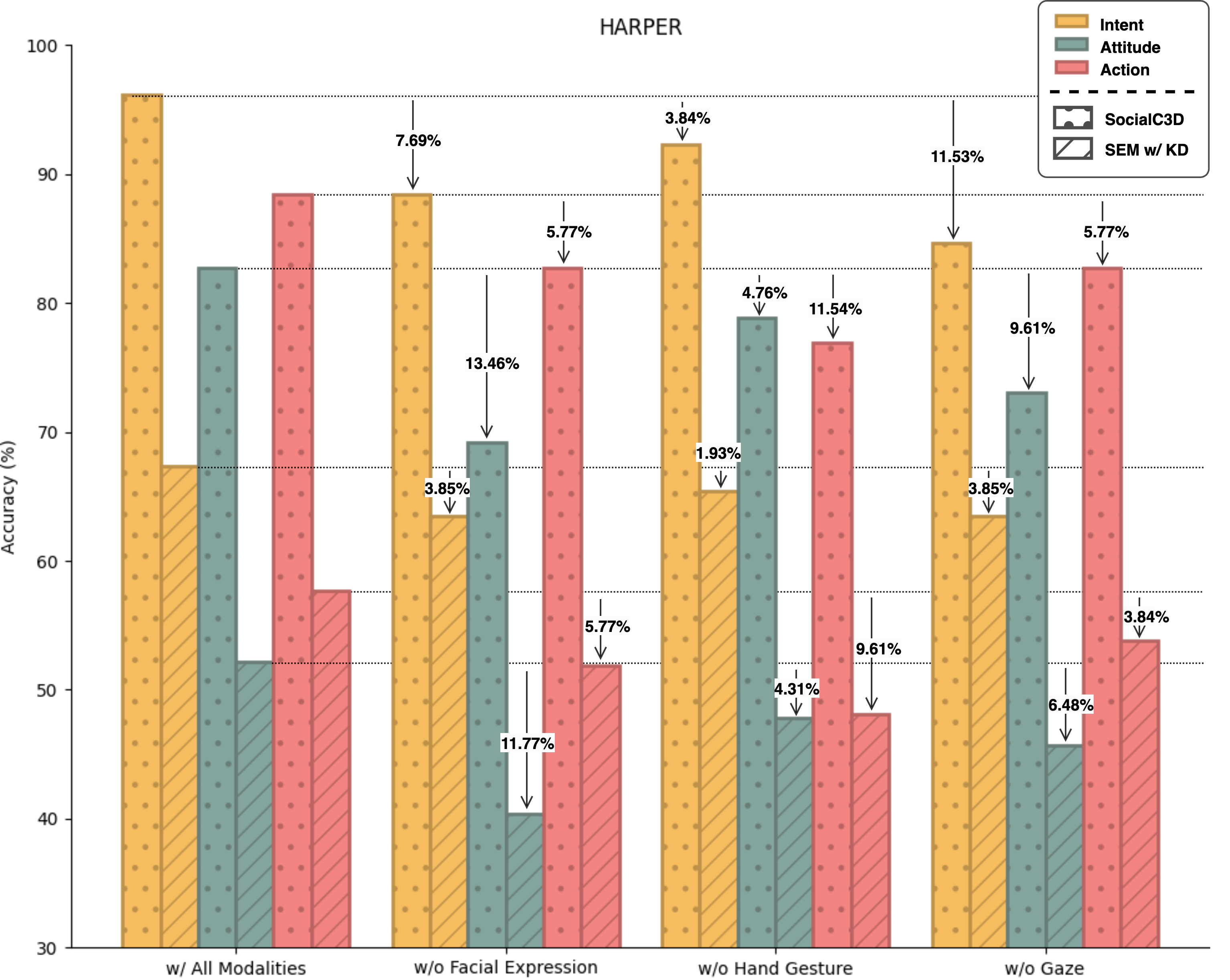}
    % \caption{HARPER}
  \end{subfigure}
  \end{minipage}
\vspace{-2mm}
\caption{Comparison of SocialC3D using different input modalities and the its impact on SocialEgoMobile (SEM). Accuracy drops due to missing modalities are labelled.}
\label{fig:modal}
\vspace{-4mm}
\end{figure*}

%% file: 4_experiments/ablation_experiments.tex
\subsection{Ablation Experiments}
\textbf{Contribution of Mutli-Modalities.}~
We incorporate body pose, raw images, face, hand gestures, and gaze as input modalities for the teacher model SocialC3D, enabling it to comprehensively capture uers' non-verbal social cues. Body pose and raw images are the foundational modalities and prior work has demonstrated their effectiveness and complementarity in understanding body movements \cite{duan2022revisiting}. In contrast, facial expressions, hand gestures, and gaze focus on localized regions of the body, providing fine-grained cues essential for capturing nuanced social signals.

Fig.\ref{fig:modal} illustrates the performance of SocialC3D and SocialEgoMobile under its supervision with full modality and with each of the three localized modalities ablated. Removing any modality degrades the overall performance of SocialC3D, but the extent of impact varies across social sub-tasks. Specifically, the \textbf{Attitude} and \textbf{Intent} tasks show performance degradation when facial information is removed. Facial orientation offers cues about a user's attention, while expressions convey emotional states that differentiate social behaviours with similar body movements in early stages, such as `throw' and `wave hands'. The \textbf{Action} task is more sensitive to the absence of hand gestures, as detailed hand pose information helps the model to distinguish between fine-grained actions, such as the differences in finger closure for `touch' and `handshake'. For the \textbf{Intent} task, the lack of gaze information leads to the greatest drop in accuracy, since gaze direction directly reflects user focus and intention. While facial orientation may offer partial substitutes, it cannot fully replicate the precision of direct gaze signals. 

For the student model, 30\% of the frames and joints are corrupted in the input. We maintain this corruption ratio to ensure that the findings remain applicable to real-world deployment scenarios. We consider this corruption setting representative, as it strikes a balance between noise and meaningful information. Through knowledge distillation, the impact of missing modalities on the teacher is transferred to the student, as shown in Fig.\ref{fig:modal}. These results confirm that the proposed knowledge distillation framework successfully transfers the integrated multimodal social knowledge from the teacher model to the unimodal student model.

\vspace{2mm}

\input{4_experiments/table_ablation_distillation}

\noindent\textbf{Distillation Methods.}~
In Table \ref{tab:ablation_kd}, we compare the performance improvements brought by different knowledge distillation methods to SocialEgoMobile. Distillation based on intermediate features outperforms soft label distillation, as the student model in our experiments must learn to integrate information of additional modalities from the teacher's social representation and compensate for the impact of noisy inputs. Learning only from the teacher’s output distributions is insufficient. Among the feature-based methods, InfoNCE \cite{oord2018representation} achieves the best performance. By maximizing mutual information between the teacher and student model's social representations, InfoNCE encourages the student to learn semantically aligned features, rather than enforcing exact numerical matching at lower levels. This is suitable for our experiment setting, where the teacher and student operate on different input modalities and the student receives noisy inputs, making low-level alignment impractical. InfoNCE promotes the alignment of high-level semantic information, leading to better noise tolerance and generalization.

\vspace{2mm}

\input{4_experiments/table_ablation_student}

\noindent\textbf{Student Model Architecture.}~
To enhance the robustness and efficiency of processing noisy pose feature sequences, we designed SocialEgoMobile with a two-layer GAT \cite{veličković2018graph} followed by a one-layer Bi-LSTM \cite{graves2005framewise}. Table \ref{tab:ablation_student} presents a comparison of different model configurations under the knowledge distillation with corrupted inputs. In the spatial dimension, a two-layer GAT achieves a good trade-off between expressiveness and stability. Fewer layers limit the receptive field, reducing the ability to capture global graph patterns, while deeper GATs tend to propagate noise more widely. Moreover, GAT's adaptive attention mechanism enables selective aggregation of useful information from neighbouring joints, offering greater resilience to noisy signals compared to standard Graph Convolutional Networks (GCN) \cite{kipf2016semi}. In the temporal dimension, a single-layer Bi-LSTM provides sufficient modelling capacity for short feature sequences, while deeper Bi-LSTMs do not result in a steady performance improvement. Compared to Temporal Convolutional Networks (TCN) \cite{lea2017temporal} and Transformers \cite{vaswani2017attention}, Bi-LSTM is better suited for sequential data with inherent temporal dependencies and benefits from its gating mechanism, which improves robustness to sparsity and temporal discontinuity. Overall, the combination of two GAT layers and one Bi-LSTM in SocialEgoMobile offers an effective balance between expressiveness, robustness, and computational efficiency.

%% file: 4_experiments/table_ablation_distillation.tex
\begin{table*}[t]
\centering
\caption{Comparison of different knowledge distillation methods when 30 \% of joints and frames are corrupted.}
\vspace{-2mm}
\renewcommand*{\arraystretch}{1.2}
\small
\resizebox{0.9\textwidth}{!}{
\begin{tabular}{l|ccc|ccc}
\toprule
\multirow{2}{*}{\textbf{Distillation Method}} &
  &
  \textbf{JPL-Social}&
  &
  &
  \textbf{HARPER}
  &\\
  &
  \textbf{Intent Acc.} &
  \textbf{Attitude Acc.} &
  \textbf{Action Acc.} &
  \textbf{Intent Acc.} &
  \textbf{Attitude Acc.} &
  \textbf{Action Acc.} \\ \midrule
w/o KD &
  80.95 &
  59.52 &
  59.52 &
  61.54 &
  28.84 &
  42.31 \\
Soft Label \cite{hinton2015distilling}&
  82.14 &
  71.43 &
  64.29 &
  61.54 &
  39.13 &
  44.23 \\
Attention Transfer \cite{zagoruyko2017paying} &
  84.52 &
  73.81 &
  69.05 &
  63.46 &
  39.13 &
  48.08 \\
FitNet \cite{romero2014fitnets} &
  84.52 &
  76.19 &
  66.67 &
  65.38 &
  43.48 &
  48.08 \\
KDGAN \cite{wang2018kdgan}&
  88.10 &
  78.57 &
  69.05 &
  65.38 &
  43.48 &
  50.00 \\
\textbf{InfoNCE} \cite{oord2018representation}&
  \textbf{88.10} &
  \textbf{80.95} &
  \textbf{75.00} &
  \textbf{67.31} &
  \textbf{52.14} &
  \textbf{57.69} \\
\bottomrule
\end{tabular}
}
\label{tab:ablation_kd}
\vspace{-2mm}
\end{table*}

%% file: 4_experiments/table_ablation_student.tex
\begin{table*}[t]
\centering
\caption{Comparison of SocialEgoMobile with different architectures when 30\% joints and frames are corrupted.}
\vspace{-2mm}
\renewcommand*{\arraystretch}{1.2}
\small
\resizebox{0.9\linewidth}{!}{
\begin{tabular}{l|ccc|ccc}
\toprule
\multirow{2}{*}{\textbf{Model Architecture}} &
  \multicolumn{3}{c}{\textbf{JPL-Social}}
  & \multicolumn{3}{c}{\textbf{HARPER}}\\
 &
  \textbf{Intent Acc.} &
  \textbf{Attitude Acc.} &
  \textbf{Action Acc.} &
  \textbf{Intent Acc.} &
  \textbf{Attitude Acc.} &
  \textbf{Action Acc.} \\ 
\midrule
\textbf{SocialEgoMobile} &
\textbf{88.10} &
80.95 &
\textbf{75.00} &
\textbf{67.31} &
\textbf{52.14} &
\textbf{57.69}\\  
\midrule
w/ 1 layers GAT &
  83.33 &
  69.05 &
  67.86 &
  63.46 &
  43.49 &
  48.09 \\
w/ 3 layers GAT &
  84.52 &
  76.19 &
  69.05 &
  65.38 &
  47.83 &
  53.85 \\
w/ 2 layer GCN \cite{kipf2016semi} &
  85.71 &
  71.43 &
  67.86 &
  59.62 &
  47.83 &
  51.92\\
\midrule
w/ 2 layers Bi-LSTM & 
  \textbf{88.10} &
  \textbf{82.14} &
  \textbf{75.00} &
  61.84 &
  \textbf{52.14} &
  51.92\\
w/ 1 layer TCN \cite{lea2017temporal}&
  83.33 &
  69.05 &
  65.48 &
  59.62 &
  39.13 &
  48.08\\ 
w/ 1 layer Transformer \cite{vaswani2017attention}&
  77.38 &
  66.67 &
  54.76 &
  55.77 &
  26.08 &
  42.31\\ 
  \bottomrule
\end{tabular}
}
\label{tab:ablation_student}
\vspace{-4mm}
\end{table*}

%% file: 4_experiments/key_takeaways.tex
\subsection{Key Takeaways}

To summarize our results, we highlight three main observations:

\begin{itemize}[leftmargin=1em, topsep=0.3em]
    \item SocialC3D surpasses SOTAs by effectively integrating body, face, hand, gaze, and image inputs for egocentric social understanding, demonstrating the value of complementary multimodal cues and the need for socially aware robots capable of comprehensively interpreting human behaviour.
    \item The knowledge distillation framework enables the student model to absorb multimodal knowledge from the teacher, improving its ability to capture meaningful information and robustness against spatial-temporal noisy data in a deployment setting.
    \item The distilled SocialEgoMobile achieves a great trade-off between performance, robustness, and computational efficiency, which is suitable for deployment on resource-constrained systems, promoting practical and scalable HRI in real-world settings.
\end{itemize}